\documentclass[runningheads]{llncs}

\usepackage{wrapfig}
\usepackage{graphicx}
\graphicspath{ {./figs/} }
\usepackage{url}
\usepackage[noadjust]{cite}  % bibtex references
\usepackage{subcaption} % automatically includes package "caption". For subfigure

\usepackage{amsmath}
\usepackage{amsfonts}  % mathbb

\usepackage[group-separator={,}]{siunitx}

\usepackage{algorithm}
\usepackage{algorithmic}

\usepackage[table]{xcolor} % add color to table cells
\usepackage{tabularx}

\usepackage{multirow} 
\usepackage{adjustbox} % adjust table size
\usepackage[misc]{ifsym} % envelope
\usepackage{xspace}
\usepackage[colorlinks=true, linkcolor=red, urlcolor=blue, citecolor=cyan, bookmarks=true, bookmarksopen, pagebackref=true]{hyperref}
\newcommand{\ac}[1]{\textcolor{magenta}{\textbf{Alvaro:} #1}\xspace}
\newcommand{\mk}[1]{\textcolor{blue}{\textbf{Murat:} #1}\xspace}

\renewcommand{\ac}[1]{}
\renewcommand{\mk}[1]{}

\title{A Systematic Evaluation of Generative Models on Tabular Transportation Data}

\titlerunning{A Systematic Evaluation of Generative Models on Transportation Data}
\author{
 Chengen Wang$^{(\textrm{\Letter})}$\inst{1} \and
 Alvaro Cardenas\inst{2} \and
 Gurcan Comert\inst{3} \and
 Murat Kantarcioglu\inst{1}
}

\authorrunning{C. Wang et al.}
\institute{
University of Texas at Dallas, Dallas TX 75080, USA\\
\email{\{chengen.wang,muratk\}@utdallas.edu} \and
University of California, Santa Cruz,  Santa Cruz CA 95064, USA
\email{alacarde@ucsc.edu}\\
\and
Benedict College, Columbia SC 29204, USA\\
\email{Gurcan.Comert@Benedict.edu}
}

\begin{document}
\maketitle              % typeset the header of the contribution

% \tableofcontents

\begin{abstract}
The sharing of large-scale transportation data is beneficial for transportation planning and policymaking. However, it also raises significant security and privacy concerns, as the data may include identifiable personal information, such as individuals' home locations. To address these concerns, synthetic data generation based on real transportation data offers a promising solution that allows privacy protection while potentially preserving data utility. Although there are various synthetic data generation techniques, they are often not tailored to the unique characteristics of transportation data, such as the inherent structure of transportation networks formed by all trips in the datasets. In this paper, we use New York City taxi data as a case study to conduct a systematic evaluation of the performance of widely used tabular data generative models. In addition to traditional metrics such as distribution similarity, coverage, and privacy preservation, we propose a novel graph-based metric tailored specifically for transportation data. This metric evaluates the similarity between real and synthetic transportation networks, providing potentially deeper insights into their structural and functional alignment. We also introduced an improved privacy metric to address the limitations of the commonly-used one. Our experimental results reveal that existing tabular data generative models often fail to perform as consistently as claimed in the literature, particularly when applied to transportation data use cases. Furthermore, our novel graph metric reveals a significant gap between synthetic and real data. This work underscores the potential need to develop generative models specifically tailored to take advantage of the unique characteristics of emerging domains, such as transportation. Our code is available at \href{https://github.com/chengenw/transportation.git}{https://github.com/chengenw/transportation.git}. 


\end{abstract}

% \begin{IEEEkeywords}
% Tabular Data Synthesis, Transportation Data, Generative Models Evaluation
% \end{IEEEkeywords}

\section{Introduction}\label{sec:intro}
% sharing transportation data: benefits and concerns
Open large-scale transportation data offer significant benefits, including enhancing transparency, improving traffic management and urban planning, and providing valuable opportunities for researchers to conduct in-depth studies that inform and support effective policymaking. However, the availability of large-scale data raises significant privacy concerns, as human mobility is highly sensitive. Transportation data may contain identifiable personal information such as individuals' home locations, and data sharing can infringe on individual privacy. For example, celebrities may be stalked with shared taxi journey data \cite{nyt_prv}.

% synthetic data generation -> privacy
To address these challenges, synthetic data generation techniques have emerged as a promising solution for data sharing, offering potentially high utility for downstream tasks while effectively mitigating the privacy concerns. 
These techniques essentially approximate the raw data distribution with machine learning models and then generate artificial data instead of sharing the raw data directly.

% Deep generative models
In recent years, deep learning-based synthetic data generation models have drawn considerable attention due to their ability to learn complex data distributions and generate realistic synthetic data. Among deep generative models, Generative Adversarial Networks (GANs) \cite{GAN}, Variational Autoencoders (VAEs) \cite{vae}, and diffusion models \cite{DDPM
% ,song2019generativeModeling
} have demonstrated remarkable capabilities in generating high-quality samples, particularly in domains such as images and text.

% general tabular data generation
Deep generative models have also been adapted for various data applications, including tabular data with structured rows and columns \cite{ctgan}. This prevalent form of data representation is widely used in diverse domains, including finance, healthcare, and e-Commerce, among others \cite{borisov2022deep_tab_survey}. Tabular data is also a common format for sharing transportation data. For instance, taxi trips can be represented in a tabular format, where each row corresponds to an individual trip, and the columns capture various properties of the trip, such as the start location, end location, and trip duration. 

Researchers have proposed a variety of models to generate synthetic tabular data, such as GAN and Variational autoencoder-based, diffusion-based generative modeling \cite{ctgan, kotelnikov2023tabddpm}. To synthesize transportation data, leveraging existing tabular data generative models provides an efficient starting point for further research. However, it is observed that current generative models are primarily designed for general tabular data synthesis, leaving several important questions unanswered:
\begin{enumerate}
    \item Transportation data possesses unique properties; for instance, the data collectively forms a transportation network. Can the existing metrics effectively capture and measure this characteristic?
    \item Tabular data generative models typically do not prioritize privacy preservation. How effectively do they perform when evaluated from a privacy preservation perspective?
    \item The datasets commonly used to evaluate these models generally do not include transportation data. Will the performance ranking of these models remain consistent when applied to transportation data?
\end{enumerate}

% contributions
In this paper, we aim to answer these questions by conducting a systematic evaluation of typical tabular data generative models \cite{sdv_paper, ctgan, ctab-gan, kim2022stasy, kotelnikov2023tabddpm} on transportation data, using New York City taxi data \cite{taxi_2015} as a case study. Our main contributions can be summarized as follows:
\begin{enumerate}
    \item We propose a novel graph-based metric to quantify the property gap between real and synthetic transportation data, leveraging the fact that transportation networks can naturally be represented as graphs.
    \item We propose an improved privacy leakage metric to investigate the privacy-preserving capabilities of these models and assess their vulnerabilities, particularly to membership inference attacks \cite{membership_MIA}.
    \item We systematically evaluate the performance of these models in the transportation domain using a comprehensive set of metrics, including downstream task utility, distribution similarity, diversity, complexity, and the two novel metrics mentioned above.
\end{enumerate}

% outline
In the remainder of this paper, we first provide a brief review of related work in Section \ref{sec:related} and present the necessary background knowledge in Section \ref{sec:background} to facilitate a better understanding of our work. Next, we introduce novel evaluation metrics in Section \ref{sec:novel_metric}. The experimental setup and results are detailed in Section \ref{sec:exp}, followed by a concluding summary in Section \ref{sec:conc}.

\section{Related Work}\label{sec:related}

\subsection{Generative Models} Generative Adversarial Networks (GANs) \cite{GAN} are among the most widely used generative models. It employs a generator and a discriminator, two competing neural networks; the generator tries to trick the discriminator to classify the fake data as real, while the discriminator tries to differentiate real and fake data.

Variational Autoencoders (VAEs) \cite{vae} are another class of generative models, which map real data to a distribution within a latent space by an encoder, then a decoder maps from the latent space to the input space.  

Diffusion models represent the latest advancement in generative models \cite{DDPM
% ,song2019generativeModeling
}. They involve a forward diffusion process and a reverse denoising process. In the forward process, noise is gradually added to the training data with increasing magnitude until the data becomes pure noise. In the reverse process, a model is trained to denoise the noisy data, effectively reconstructing the clean data and learning the underlying data distribution.

\subsection{Tabular Data Generation}
Tabular data pose unique challenges for synthetic data generation. Unlike image data, tabular data often consist of a mix of continuous and discrete variables. Moreover, values in the discrete columns frequently exhibit imbalanced distributions, adding an additional layer of complexity to the generation process. \cite{ctgan} proposes a conditional tabular GAN (CTGAN) to address these challenges. CTGAN employs two distinct sampling approaches to handle discrete and continuous variables in the training data. For discrete variables, it first randomly selects a discrete column, then samples rows based on the logarithm frequency of categorical values in that column. The sampled categorical values will serve as conditional inputs to GAN. For continuous variables, it estimates the number of modes for each column with variational Gaussian mixture models \cite{bishop2006pattern} and samples by modes and normalizes the values. \cite{ctgan} also proposed tabular VAE (TVAE) by adapting VAE to tabular data.

\cite{ctab-gan} makes improvements upon CTGAN motivated by several observations: \emph{Within one variable} of the tabular data there may be mixed continuous and categorical data types, and its distribution may be skewed and have a long tail. The authors address these issues by proposing mode-value pair for mixed data types, logarithmic transformation for variables with long tail distribution, and an additional continuous mode as the conditional input to GAN.

\cite{kotelnikov2023tabddpm} proposed TabDDPM by adapting diffusion models to the tabular data domain, employing Gaussian diffusion models for continuous variables and multinomial diffusion models for categorical variables \cite{hoogeboom2021argmax}. \cite{kim2022stasy} proposed the STaSy model by directly adapting score-based generative modeling \cite{song2021scorebased_sde} to the tabular data domain.

While these techniques have shown promise in tabular data, to the best of our knowledge, they have not been evaluated in the context of transportation data with their unique characteristics. 
In this work, we evaluate the aforementioned tabular synthetic data generation techniques within the context of a transportation data use case.

\section{Background}\label{sec:background}
\subsection{Diffusion Models for Tabular Data Generation}
As diffusion models represent the latest advancements in generative modeling and are less widely known compared to GANs or VAEs, we provide a brief overview of these models in this section to facilitate understanding.

Let the training data be $x_0\sim q(x_0)$. In the forward process, Gaussian noise is added to the clean data, and the diffusion process in DDPM \cite{DDPM} is formulated as
\begin{equation}
    q(x_t|x_{t-1}):=\mathcal{N}(x_t;\sqrt{1-\beta_t}x_{t-1},\beta_t\mathbf{I})
\end{equation}
where $\beta_t$ is the variance schedule for the Gaussian noise. Based on the Bayesian theorem, we can calculate the reverse process $q(x_{t-1}|x_t, x_0)$. We use a neural network to represent the denoising process as
\begin{equation}
    p_\theta(x_{t-1}|x_t)=\mathcal{N}(x_{t-1};\boldsymbol{\mu}_\theta(x_t,t),\boldsymbol{\Sigma}_\theta(x_t,t)).
\end{equation}
After setting $\boldsymbol{\Sigma}_\theta(x_t,t)$ to untrained time-dependent constants, the training loss is simplified to learn the added Gaussian noise $\epsilon$ as follows:
\begin{equation}
    \mathbb{E}_{x_0,t,\epsilon\sim\mathcal{N}(0,\boldsymbol{I})}\left[w(t)\|\epsilon-\epsilon_\theta(x_t,t)\|^2\right],
\end{equation}
where $w(t)$ is a weight function.

% This Gaussian diffusion process defined above works on continuous data space. 
The generative model TabDDPM \cite{kotelnikov2023tabddpm} mentioned in Section \ref{sec:related} leverages multinomial diffusion models to generate categorical data. TabDDPM's forward process corrupts the categorical data by adding uniform noise over $K$ classes as follows \cite{kotelnikov2023tabddpm}:
\begin{equation}
\begin{aligned}
    q(x_t|x_{t-1}):=\mathrm{Cat}(x_t;(1-\beta_t)x_{t-1}+\frac{\beta_t}{K}),\\
    q(x_T):=\mathrm{Cat}(x_T;\frac{1}{K}),
\end{aligned}
\end{equation}
where $x_t$ is a one-hot encoded categorical variable with $K$ values.

\cite{song2021scorebased_sde} generalizes the diffusion models to the continuous-time domain. The forward process can be modeled as the solution to a stochastic differential equation (SDE):
\begin{equation}
    \mathrm{d}x=\boldsymbol{f}(x,t)\mathrm{d}t + g(t)\mathrm{d}\boldsymbol{w},
\end{equation}
where $\boldsymbol{w}$ is the standard Wiener process, $\boldsymbol{f}(\cdot,t)$ is the drift coefficient, and $g(t)$ is the diffusion coefficient. The reverse process is given by the reverse-time SDE:
\begin{equation}
    \mathrm{d}x=\left[\boldsymbol{f}(x,t)-g(t)^2\nabla_x\log p_t(x)\right]\mathrm{d}t + g(t)\mathrm{d}\bar{\boldsymbol{w}},
\end{equation}
where $\bar{\boldsymbol{w}}$ is the standard Wiener process in reverse time, and $\nabla_x\log p_t(x)$ is the score of the probability density $p_t(x)$. The STaSy paper \cite{kim2022stasy} mentioned in Section \ref{sec:related} is proposed based on the SDE generative modeling.

\subsection{Evaluation Metrics for Generative Models} \label{sec:metrics}
The quality of synthetic data is evaluated based on its utility and its ability to preserve privacy. However, higher privacy protection can potentially reduce a model's utility. Conversely, higher utility (i.e., synthetic data that closely resembles real data) may increase the privacy risks. Balancing utility and privacy in synthetic data generation is challenging and remains an active research area \cite{groundhog}. 

In this context, utility can be assessed through various measures, including downstream task performance (e.g., using synthetic data for machine learning tasks), statistical similarity between the original data and the generated synthetic data, such as the Wasserstein distance \cite{wasserstein-distance}, and diversity metrics, such as coverage \cite{coverage}. Privacy preservation can be evaluated by measuring the distance between a synthetic data point and its nearest neighbor in the real data, which provides an indication of how closely the synthetic data mirrors individual records in the original dataset \cite{hilprecht2019_gan_attack}. Below, we present the mathematical formalization of these measures.

\subsubsection{Downstream Task Performance} To evaluate the utility of the generated synthetic data, we employ a selected downstream task. In the context of taxi ride information, one critical piece of information is the total cost of the ride. The key question, therefore, is how effectively the synthetic data can be used to train a machine learning model capable of accurately predicting the total cost of a taxi ride. More specifically, we generate synthetic data with the trained generative model, train a prediction model ``Gradient Boosting for Regression" \cite{gradient_boosting} with the synthetic data, then we predict the ``total amount" (i.e., the total amount paid for the taxi ride) with the training data and synthetic data respectively. The performance is represented by coefficient of determination \cite{coefficient_determination}
\begin{equation}
   R^2=(1-\frac{u}{v}),
\end{equation}
where
\begin{equation}
   u=\sum_i(y-\hat{y})^2, v=\sum_i(y-\bar{y})^2,
\end{equation}
where $y$ is the true value, $\hat{y}$ is the predicted value and $\bar{y}$ is the mean of the true values. 
The best possible value of $R^2$ is 1.

\subsubsection{Similarity}
%One metric is similarity score \cite{sdv_score}, which is $1-\delta_{KS}$ for numerical value, and $1-\delta_{TVD}$ for discrete values, where $\delta_{KS}$ is KS statistics and $\delta_{TVD}$ is Total Variation Distance. The range of the score value is [0, 1]: the higher the score, the more similar of two distributions.

We use Wasserstein distance to measure the similarity between two distributions (i.e., real vs synthetic data), which can be represented as \cite{WGAN}:
%\mk{Do you need the brackets \textbackslash[ \textbackslash ] in the formula below.} % notation convention: E[]
\begin{equation}
    W(\mathbb{P}_r,\mathbb{P}_g)=\inf_{\gamma\in\Pi(\mathbb{P}_r,\mathbb{P}_g))}\mathbb{E}_{(x,y)\in\gamma}[\|(x-y)\|]
\end{equation}
where $\Pi(\mathbb{P}_r,\mathbb{P}_g)$ is the set of all joint distributions $\gamma(x,y)$ whose marginals are $\mathbb{P}_r$ and $\mathbb{P}_g$, respectively.

\subsubsection{Diversity} We use coverage \cite{coverage} to measure the diversity of a distribution, enabling us to assess whether mode collapse \cite{GAN} has occurred. Coverage is calculated as the percentage of real sample hyperspheres which contain a generated sample. The real sample hypersphere is calculated with its $K^{th}$ nearest neighbor. It is found to be more robust than the metric recall \cite{improved_precision_recall, thompson2022evaluation}.

\subsubsection{Privacy Measure} \label{sec:dcr}
The distance of a synthetic data point to its closest real data neighbor (DCR) serves as a metric for evaluating privacy preservation in synthetic data generation \cite{dcr_renmin, ctab-gan}. This ensures that synthetic records are not overly similar to individual records in the original dataset, thereby reducing the risk of privacy breaches.
%\mk{the next sentence is not clear need to expand on it.}
This metric is also closely related to membership inference attacks \cite{membership_MIA}, where a distance-based metric \cite{hilprecht2019_gan_attack} is utilized to determine whether a data point was included in the training dataset of the model under attack. We explore this connection in greater detail in Section \ref{sec:rDCR}.

\section{The Novel Evaluation Metrics}\label{sec:novel_metric}
In this section, to evaluate the generated synthetic data for transportation applications, we propose two novel metrics. 
%\mk{Please check whether the next claim is correct:} % there is a similar one to rDCR, but not the same
%MK2: I added transportation to make it even more different.
To the best of our knowledge, these metrics have not been previously used in the context of evaluating synthetic tabular transportation data generation. 

\subsection{Graph Similarity Metric for Transportation Network Data}
Transportation data, when viewed collectively, forms a transportation network. This network can be effectively represented as a graph, capturing the overall transportation trends and relationships within the data. For instance, the pickup and drop-off locations in the NYC taxi dataset correspond to different zones within the city. These zones can be represented as nodes in a graph, providing a structured way to model the transportation network. In other words, each trip between two zones can be represented as an edge in the transportation graph. Let the number of trips between two zones $i$ and $j$ be $n_{ij}$, where $i,j\in\mathbb{N}^+$, then the total number of trips $N=\sum_{i,j}n_{ij}$. Let the fraction of edges between two nodes $i$ and $j$ of the transportation graph $G$ be $p_G(i,j)=\frac{n_{ij}}{N}$. Clearly, $\sum_{i,j}p_G(i,j)=1$, which means the fraction of edges $p_G$ represents a distribution.

%\mk{I change the notation to make it graph centric. Please check.} 
% revised
We can construct a transportation graph from the real transportation data, denoted as $G_r$, and another graph, $G_s$, from the generated synthetic transportation data. We can measure the similarity between the real transportation graphs $G_r$ and the synthetic ones $G_s$ by calculating the similarity score $S_G$ between the two graphs as follows:
\begin{equation}
\begin{aligned} % break an equation into multiple lines, and numbering the equation once
   S_G(G_r, G_s) = 1 - \delta(p_{G_r}, p_{G_s}), \\
   s.t. \quad \delta(p_{G_r}, p_{G_s}) = \frac{1}{2}\sum_{i,j}|p_{G_r}(i,j) - p_{G_s}(i,j)|,
    % S_G(G_r,G_s) = 1 - \delta(G_r, G_s), \\
    % s.t. \quad \delta(G_r, G_s) = \frac{1}{2}\sum_{i,j}|G_r(i,j) - G_s(i,j)|,
\end{aligned}    
\end{equation}
where $p_{G_r}$ and $p_{G_s}$ represent edge number distributions for the real and synthetic graphs $G_r, G_s$ respectively, and $\delta(p_{G_r},p_{G_s})$ is the total variation distance.

\subsection{Distance to Closest Record Ratio as Privacy Leakage Metric} \label{sec:rDCR}
The success of membership inference attacks relies on the observation that models tend to overfit their training data \cite{membership_MIA}. Consequently, the distance between \emph{training} data and synthetic data is smaller than the distance between \emph{testing} data and synthetic data. This phenomenon is also evidenced by the fact that training data loss is typically smaller than testing data loss. Based on this observation, relying solely on the distance between real data and synthetic data---without differentiating two types of real data, training data and testing data--as commonly done in previous literature \cite{ctab-gan, dcr_RSD}, may be insufficient to reliably assess the risk of privacy leakage.

%\mk{We need to further justify why this metric is more robust ???}
We therefore propose a more robust metric that uses two distances instead of one, comparing the two distances by calculating their ratio. Specifically, we set aside holdout \emph{testing} data beside the \emph{training} data. Let the distance of training data to the closest synthetic data be $d_\alpha(r,s)$, and the distance of holdout testing data to the closest synthetic data be $d_\alpha(h,s)$, where $\alpha$ is the percentile of all the closest distance values, respectively, after ranking each set of distances in increasing order. The Distance to Closest Record Ratio (rDCR) is defined as
% follows:
% \begin{equation} \label{eq:ratio}
%     r_{DCR}=\frac{d_\alpha(r,s)}{d_\alpha(h,s)}.
% \end{equation}
$r_{DCR}=\frac{d_\alpha(r,s)}{d_\alpha(h,s)}.$

%\mk{Next sentences require justification. I would like to check the update once you are done. This is very critical !!!}
When $r_{DCR} < 1$, the distance between the training data and synthetic data is smaller than the distance between testing and synthetic data. This indicates overfitting, making the model vulnerable to membership inference attacks. 
\mk{I modified this more. Please check.} % ok
A smaller \emph{ratio} $r_{DCR}$ indicates greater overfitting of the model to the training data, making the model more vulnerable to a potential membership inference attack. On the other hand, if $r_{DCR} > 1$, a small \emph{distance} of $d_\alpha(r,s)$ alone may not be sufficient to demonstrate the vulnerability of a model to distanced-based membership inference attacks.
\mk{what about other MIA attacks?} % here is the limitation of the distance based metric

The metric from \cite{platzer2021holdout} bears a resemblance to the rDCR metric described in this work, but there are significant differences. Our metric focuses on the distance to the closest synthetic data for each real data as it is designed to analyze the privacy leakage for the real data. In contrast, their metric measures the closest distance to real data for each synthetic data. Additionally, we incorporate a percentile-based approach to assess privacy leakage, recognizing that typically only \emph{a small percentage} of the training data is vulnerable to membership inference attacks~\cite{carlini2022membership_first_principles}. By examining the percentile of data at risk of privacy leakage, this approach provides a more refined method for assessing privacy risks.

% \mk{Some of the columns miss the bold font. Please check and fix.} % values in those columns provided for reference purposes only. this is mentioned in the main text.

\section{Experiments}\label{sec:exp}

\subsection{Datasets} \label{sec:dataset}
% sampling, data size
In this work, we use New York City taxi trip data \cite{nyc_taxi} as the experimental dataset. 
%\mk{I am worried that the reviewers say why just one dataset. I tried to answer it. Please check and add citations.}
This dataset is widely utilized in transportation research \cite{
% riascos2020networks_nyc_taxi, 
correa2017exploring_nyc_taxi, 
mauro2022generating_nyc_taxi} and is one of the largest publicly available datasets in the domain. Its extensive size and accessibility make it a valuable resource for studies in this field. It is represented in a tabular data format, with each row containing detailed information about an individual taxi trip.
% Given these characteristics, we believe that using this dataset alone is sufficient for our evaluation, as it provides a robust and comprehensive basis for assessing the performance and utility of synthetic data generation techniques in the transportation domain. 
%\mk{Not sure we want say the next sentence but I added as a place holder. Or We can mention this in the conclusion too ??} % not sure too. I add one more sentence above. the first cited paper also uses this data alone
% However, while the dataset's scale and relevance justify its selection, incorporating additional datasets in future work could further validate the generalizability of the findings. 

More specifically, we use ``2015 Green Taxi Trip Data" \cite{taxi_2015}. It has $19.2$ million rows and each row has $21$ columns. Each row represents a single trip in a green taxi, and the column fields include location and time for both pickup and drop-off, trip distance, itemized fares, payment type, tax and passenger count etc.

We pre-process the data by dropping columns `Ehail\_fee' due to too many `NaN' values, changing each pickup/drop-off column `datetime' into two columns `weekday' and `time`. The final data has 22 columns with 8 categorical variables, 2 integer variables and 12 float-type numerical variables. This transportation dataset is more complicated than the tabular datasets usually used in the previous tabular data generative model papers, due to its larger data size, mixed data type and higher dimensionality. In the experiments, we randomly sample a subset of the data: the test dataset size is \num{20000}, and the training data size is \num{40000} by default, unless specified otherwise.

%\mk{Need a sentence to clarify how 2015 and 2019 data used jointly ?} % revised
\mk{Please check my modification.} % ok
The proposed graph metric requires knowledge of the zones into which the pickup and drop-off longitude and latitude coordinates fall for each trip. To fulfill this requirement, we utilize the New York City Green Taxi Trip Records from March 2019. A key distinction of this dataset, compared to the aforementioned one, is its less granular nature: locations are represented by zones rather than precise longitude and latitude coordinates. Notably, the zone variable is categorical, unlike longitude and latitude, which are numerical. This difference makes the dataset particularly well-suited for zone-based transportation metrics.

\subsection{Generative Models Used for Evaluation} \label{sec:gm_eval}
The generative models evaluated in this paper include Gaussian Copula \cite{sdv_paper
% , sdv_doc
}, CTGAN and TVAE \cite{ctgan}, CTABGAN \cite{ctab-gan} and two diffusion-based tabular data generative models: TabDDPM \cite{kotelnikov2023tabddpm} and STaSy \cite{kim2022stasy}. We evaluate these models using various metrics including utility, similarity, diversity and privacy leakage as detailed in Section \ref{sec:metrics} and \ref{sec:novel_metric}.

\subsection{Experimental Setup}
During the experiments, for each method, three models are trained, and five times of sampling are conducted for each trained model. We limit the sample size to \num{20000} for each sampling iteration due to the high memory demands of the following Wasserstein distance calculations. The results are reported as the mean and standard deviation across a total of $15$ sampling iterations.

\begin{table*}[h!t]
\vspace{-5mm}
\caption{In the downstream task performance testing, the model predicts the "total amount". The $R^2$ values are multiplied by 100. ``dwn\_tr\_syn" refers to training on training data and predicting on synthetic data, and similarly for the other columns, as mentioned in the main text.}
\label{tab:downstream}
% \begin{adjustbox}{width=0.8\textwidth, center} % IEEE
\begin{adjustbox}{width=\textwidth, center} 

\begin{tabular}{|l|l|l|l|l|l|l|}
\hline
model          & dwn\_tr\_tr  & dwn\_tr\_syn          & dwn\_tr\_te  & dwn\_syn\_syn & dwn\_syn\_tr          & dwn\_syn\_te          \\ \hline
GaussianCopula & 99.93 (0.00) & 75.01 (2.90)          & 98.80 (0.33) & 99.33 (0.01)  & 77.55 (1.20)          & 80.64 (0.23)          \\ \hline
CTGAN          & 99.93 (0.01) & 54.34 (1.19)          & 98.86 (0.42) & 69.41 (1.40)  & 78.05 (2.29)          & 80.57 (2.21)          \\ \hline
TVAE           & 99.93 (0.01) & \textbf{82.42 (2.00)} & 98.84 (0.30) & 90.03 (0.76)  & 72.32 (2.41)          & 74.27 (2.83)          \\ \hline
CTABGAN        & 99.93 (0.00) & 2.34 (23.57)          & 98.75 (0.35) & 46.28 (3.09)  & 54.71 (20.50)         & 53.93 (22.83)         \\ \hline
STaSy          & 99.93 (0.00) & 54.93 (6.97)          & 98.68 (0.41) & 76.87 (3.78)  & 88.20 (4.45)          & 88.35 (4.74)          \\ \hline
TabDDPM        & 99.93 (0.00) & 60.26 (20.20)         & 98.92 (0.29) & 89.38 (4.77)  & \textbf{94.58 (1.38)} & \textbf{94.69 (1.55)} \\ \hline
\end{tabular}

\end{adjustbox}
\vspace{-5mm}
\end{table*}

\subsection{Experimental Results}
We report the results in separate tables, with bold fonts to highlight the best performance values, except for columns with values provided for reference purposes. All columns without synthetic data involved are for reference purposes.

\subsubsection{Downstream Task Performance}
As described in Section \ref{sec:metrics}, the model predicts the total amount for a given trip based on the other trip information. The results are reported in Table~\ref{tab:downstream}, where ``model" is the model name, ``dwn\_tr\_tr" means training on training data and predicting on training data, ``dwn\_tr\_syn" means training on training data and predicting on synthetic data, ``dwn\_tr\_te" means training on training data and predicting on testing data, provided for reference purpose, ``dwn\_syn\_syn" means training on synthetic data and predicting on synthetic data, ``dwn\_syn\_tr" means training on synthetic data and predicting on training data, and ``dwn\_syn\_te"  means training on synthetic data and predicting on testing data. The three columns ``dwn\_tr\_syn", ``dwn\_syn\_tr" and ``dwn\_syn\_te"  demonstrate the performance of these models. The performance of the models trained on synthetic data is particularly critical, as in real-world applications of data synthesis, only the synthetic data is typically made public for downstream tasks. The $R^2$ values in the table are multiplied by 100. Our results demonstrate that TabDDPM achieves the best downstream task performance among the evaluated methods.

\mk{Again some columns do not have bold results in table 2. Please check and adapt.} % mentioned in main text, some columns for reference purpose only
\begin{table}[h!t]
\vspace{-5mm}
\caption{The Wasserstein distances.}
\label{tab:similarity}
% \begin{adjustbox}{width=0.49\textwidth, center} % IEEE
\begin{adjustbox}{width=0.7\textwidth, center} 

\begin{tabular}{|l|l|l|l|}
\hline
model          & w1\_tr\_te      & w1\_tr\_syn              & w1\_te\_syn              \\ \hline
GaussianCopula & 0.1365 (0.0062) & 1.0357 (0.0935)          & 1.0340 (0.0961)          \\ \hline
CTGAN          & 0.1403 (0.0129) & 0.7078 (0.0841)          & 0.6991 (0.0803)          \\ \hline
TVAE           & 0.1262 (0.0046) & 0.9093 (0.1090)          & 0.9057 (0.1142)          \\ \hline
CTABGAN        & 0.1436 (0.0092) & \textbf{0.4260 (0.0193)} & \textbf{0.4306 (0.0216)} \\ \hline
STaSy          & 0.1243 (0.0035) & 1.0418 (0.0437)          & 1.0328 (0.0357)          \\ \hline
TabDDPM        & 0.1230 (0.0025) & \textbf{0.4421 (0.0326)} & \textbf{0.4442 (0.0331)} \\ \hline
\end{tabular}

\end{adjustbox}
\vspace{-5mm}
\end{table}

\subsubsection{Statistical Similarity}
We report the experimental results for the Wasserstein distances in Table~\ref{tab:similarity}, where ``w1\_tr\_te" is the Wasserstein distance between the training data and the testing data, provided for reference purpose, ``w1\_tr\_syn" is the Wasserstein distance between the training data and the synthetic data, and ``w1\_te\_syn" is the Wasserstein distance between the testing data and the synthetic data. The results demonstrate that CTABGAN and TabDDPM have the best performance among all the models.

\begin{table}[h!]
\vspace{-5mm}
\caption{The graph similarity score, where the original values are multiplied by 100.}
\label{tab:graph_metric}
% \begin{adjustbox}{width=0.45\textwidth, center} % IEEE
\begin{adjustbox}{width=0.6\textwidth, center} 

\begin{tabular}{|l|l|l|l|}
\hline
model          & G\_tr\_te    & G\_tr\_syn            & G\_te\_syn            \\ \hline
GaussianCopula & 73.17 (0.00) & 28.56 (0.87)          & 27.21 (0.31)          \\ \hline
CTGAN          & 73.17 (0.00) & 25.87 (0.72)          & 24.69 (0.38)          \\ \hline
TVAE           & 73.17 (0.00) & \textbf{33.21 (2.41)} & \textbf{32.67 (2.04)} \\ \hline
CTABGAN        & 73.17 (0.00) & \textbf{32.12 (6.09)} & \textbf{29.96 (6.06)} \\ \hline
STaSy          &   73.17 (0.00)          &    N/A                     &       N/A                  \\ \hline
TabDDPM        & 73.17 (0.00) & 11.56 (2.84)          & 11.17 (2.80)          \\ \hline
\end{tabular}

\end{adjustbox}
\vspace{-5mm}
\end{table}
\subsubsection{Graph Similarity Metric} \label{sec:exp_graph-metric}
We report the graph similarity results in Table~\ref{tab:graph_metric}, where ``G\_tr\_te" is the graph similarity between the training data and the testing data, provided for reference purpose, ``G\_tr\_syn" is the graph similarity between the training data and the synthetic data, ``G\_te\_syn" is the graph similarity score between the testing data and the synthetic data. The similarity values are
multiplied by 100 in the table. 
%\mk{where this reference value comes from? Please explain. }
The results indicate that all models exhibit a significant performance gap compared to the reference value $73.17$ given in column ``G\_tr\_te". The TabDDPM model shows particularly low graph metric values. Further investigation shows that TabDDPM suffers severe mode collapse. We speculate that this issue may arise from its difficulty in handling categorical variables with hundreds of classes, such as the ``zone" variable, or it may require significant additional hyperparameter tuning. 
\mk{Also, here I think you are changing the dataset for evaluating graph metric ?? not on the original 2015 data ?? We need to clarify this.} % done
Note that here we use the dataset with zone-based locations, as mentioned in Section \ref{sec:dataset}.
%\mk{Need to get the STaSy results ??}
The results for the STaSy model are unavailable due to out-of-memory related issues, likely caused by challenges in achieving convergence.

\begin{table}[h!t]
\vspace{-5mm}
\caption{The Coverage in the table is reported as the percentage of the coverage.}
\label{tab:diversity}
% \begin{adjustbox}{width=0.45\textwidth, center} % IEEE
\begin{adjustbox}{width=0.6\textwidth, center} 

\begin{tabular}{|l|l|l|l|}
\hline
model          & cov\_tr\_te  & cov\_tr\_syn          & cov\_te\_syn          \\ \hline
GaussianCopula & 74.60 (0.24) & 0.62 (0.12)           & 0.65 (0.13)           \\ \hline
CTGAN          & 74.88 (0.17) & 19.95 (1.40)          & 19.52 (1.33)          \\ \hline
TVAE           & 74.78 (0.17) & 13.94 (0.48)          & 13.79 (0.43)          \\ \hline
CTABGAN        & 74.83 (0.15) & 2.96 (0.13)           & 3.02 (0.11)           \\ \hline
STaSy          & 74.80 (0.12) & 2.34 (0.34)           & 2.40 (0.32)           \\ \hline
TabDDPM        & 75.01 (0.44) & \textbf{68.69 (0.61)} & \textbf{67.92 (0.69)} \\ \hline
\end{tabular}

\end{adjustbox}
\vspace{-5mm}
\end{table}
\subsubsection{Diversity}
We report the percentage of the coverage in Table~\ref{tab:diversity}, where ``cov\_tr\_te" is the coverage of the training data by the testing data, provided for reference purpose, ``cov\_tr\_syn" is the coverage of the training data by synthetic data and ``cov\_te\_syn" is the coverage of the testing data by the synthetic data. The coverage values are multiplied by 100 in the table. Clearly TabDDPM has the best performance. The results also reveal that all other models suffer mode dropping or collapse \cite{thompson2022evaluation}, as shown by their small coverage values.

\begin{figure}[h]
     \centering
     \includegraphics[width=0.8\textwidth]{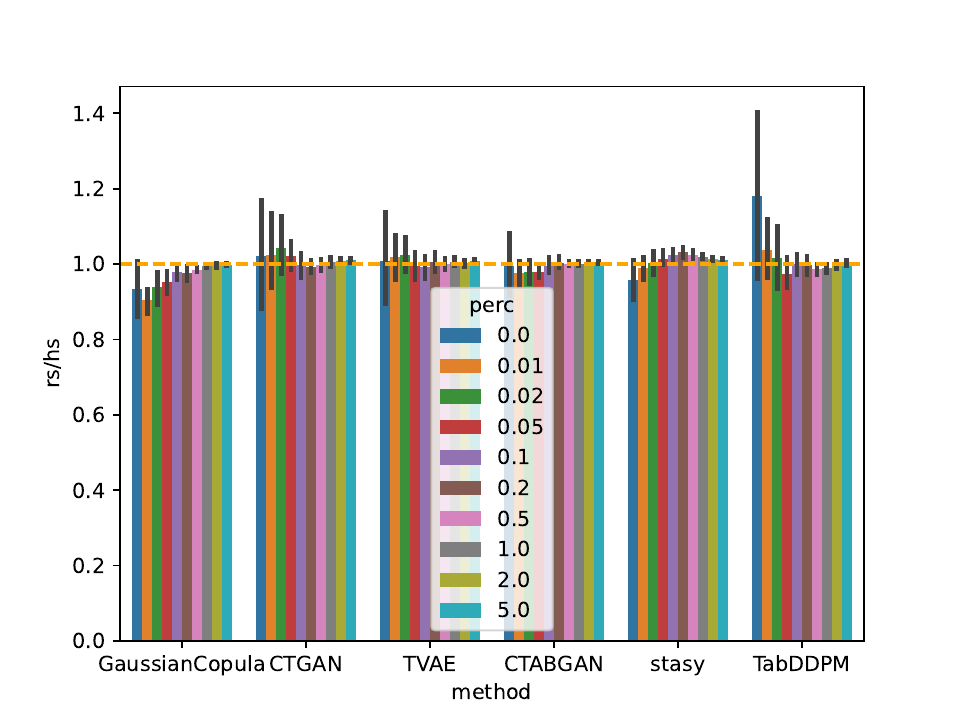}
     \caption{Privacy leakage assessment based on rDCR, i.e., DCR ratio $rs/hs$.}
     \label{fig:ratio}
 \end{figure}

 % \begin{wrapfigure}{r}{0.45\linewidth}
 %    \vspace{-10mm}
 %     \includegraphics[width=\linewidth]{figs/ratio_4k.pdf}
 %     \vspace{-6mm}
 %     \caption{Privacy leakage assessment based on DCR ratio.}
 %     \vspace{-14mm}
 %     \label{fig:ratio}
 % \end{wrapfigure}
\begin{table*}[h!]
\vspace{-5mm}
\caption{The 5\% quantile of the distances to the nearest neighbor.}
\label{tab:privacy}
% \begin{adjustbox}{width=0.8\textwidth, center} % IEEE
\begin{adjustbox}{width=\textwidth, center} 

\begin{tabular}{|l|l|l|l|l|l|c|}
\hline
model          & dcr\_rs                & dcr\_hs       & rDCR         & dcr\_rr       & dcr\_ss       & percentile \\ \hline
GaussianCopula & \textbf{0.118 (0.024)} & 0.118 (0.024) & 1.001 (0.005) & 0.003 (0.000) & 0.069 (0.005) & 5          \\ \hline
CTGAN          & 0.012 (0.002)          & 0.012 (0.002) & 1.016 (0.009) & 0.004 (0.001) & 0.014 (0.002) & 5          \\ \hline
TVAE           & 0.007 (0.000)          & 0.007 (0.000) & 1.002 (0.010) & 0.003 (0.000) & 0.006 (0.000) & 5          \\ \hline
CTABGAN        & 0.027 (0.003)          & 0.027 (0.002) & 1.004 (0.004) & 0.004 (0.001) & 0.029 (0.003) & 5          \\ \hline
STaSy          & 0.026 (0.001)          & 0.026 (0.001) & 1.010 (0.005) & 0.003 (0.000) & 0.020 (0.001) & 5          \\ \hline
TabDDPM        & 0.003 (0.000)          & 0.003 (0.000) & 1.006 (0.011) & 0.003 (0.000) & 0.003 (0.000) & 5          \\ \hline
\end{tabular}

\end{adjustbox}
\vspace{-5mm}
\end{table*}

\subsubsection{Privacy Leakage Metric}
We report the privacy leakage metric results in Table~\ref{tab:privacy}, where ``dcr\_rs" is the distance to the closest synthetic record from each real training data, ``dcr\_hs" is the DCR from holdout data to synthetic data, ``dcr\_rr" the DCR within real data, ``dcr\_ss" is the DCR within synthetic data, ``rDCR" is the ratio $\frac{d_\alpha(r,s)}{d_\alpha(h,s)}$, and ``percentile" is the $\alpha$ of the DCRs, as described in Section \ref{sec:rDCR}.

Based solely on ``dcr\_rs", as commonly used in previous literature \cite{ctab-gan}, the Gaussian Copula model has the best privacy preservation, while the TabDDPM model has the worst. 
%\mk{Please clarify the next sentence more:}
However, the results also demonstrate that none of the ``dcr\_rs" is smaller than ``dcr\_rr", which implies that actually there is possibly no privacy leakage as the synthetic data is far away from the real data.
%\mk{Remind the discussion here shortly.}
Just as we discussed in Section \ref{sec:rDCR}, ``dcr\_rs" alone may be insufficient to assess the risk of privacy leakage. 

With the proposed DCR ratio metric, we calculate the $rDCR$ for different percentile $\alpha$ values, and present the results in Figure \ref{fig:ratio}. As shown in the figure, contrary to the above conclusion, the Gaussian Copula model is vulnerable to membership inference attacks at very small values of $\alpha$, where its DCR ratio is smaller than $1$, while all the other models appear to be robust against distance-based membership inference attacks, as their DCR ratio remains approximately $1$. This finding highlights the advantage of the percentile-based ratio metric.

\begin{figure}[h]
     \centering
     \includegraphics[width=0.8\textwidth]{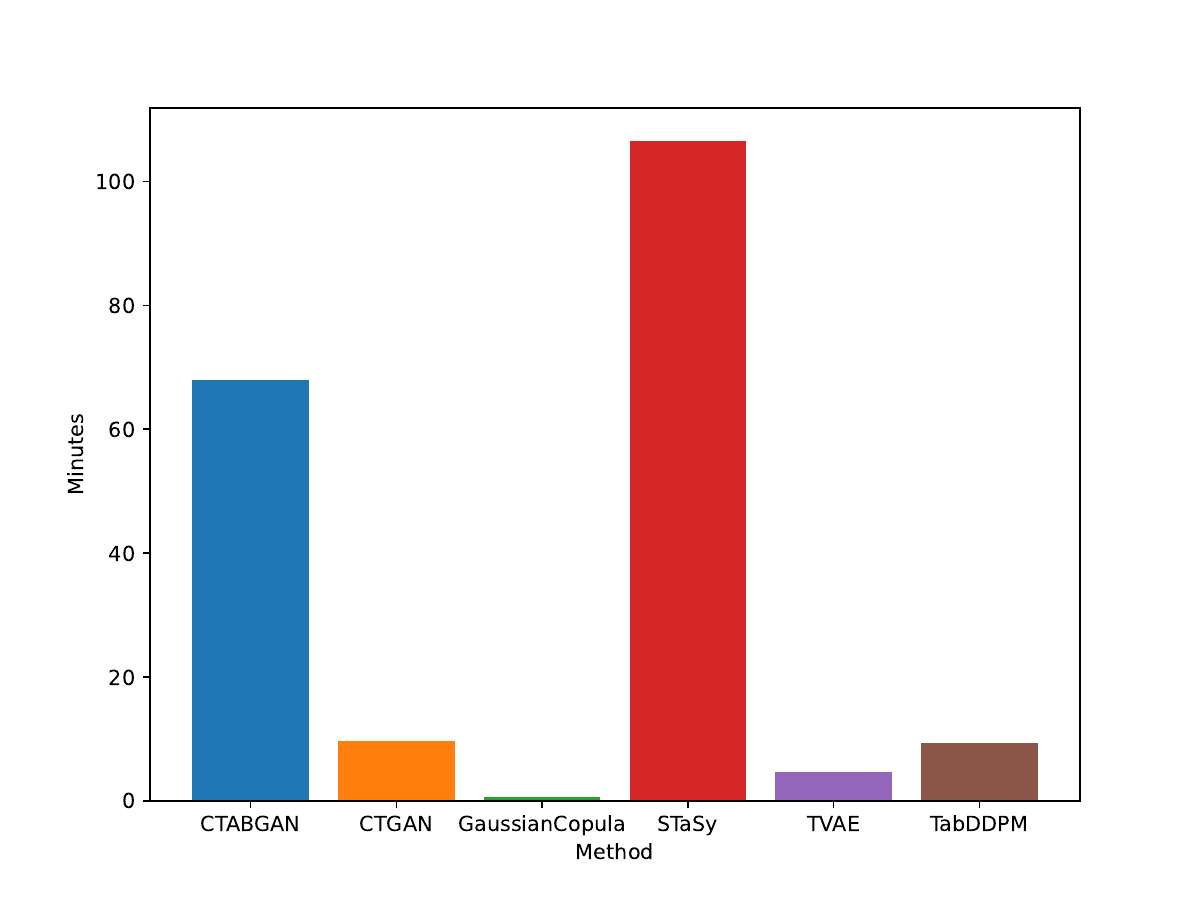}
     \caption{The complexity of the generative models in terms of running time in minutes. All the models are tested on \num{40000} samples.}
     \label{fig:speed}
 \end{figure}

\subsubsection{Complexity}
%\mk{you need the give the system settings so that we can understand. Ie, we ran using Intel X, Nvidia Y memory z etc.}
We evaluate the complexity of these models by comparing their running times. Note that the running time for diffusion-based models includes both the training and sampling time, whereas for other models, it consists only of the training time. The results are presented in Figure \ref{fig:speed}. The running times are reported in minutes, obtained from a machine with Intel(R) Core(TM) i9-9900X CPU, 64G memory, and GeForce RTX 2080. The results demonstrate that CTABGAN and STaSy models have much higher time complexity than others. Although the Gaussian Copula model has the fast training speed, its performance is not satisfactory, especially as evidenced by its minimum coverage values, severe mode collapse and possible privacy leakage.
%\mk{what about privacy ??}
The results indicate that TabDDPM achieves the best balance between speed and performance.

\section{Conclusion}\label{sec:conc}
In this paper, we conduct a systematic evaluation of generative models for synthetic tabular transportation data generation. The evaluation is conducted based on a variety of metrics including downstream tasks performance, distribution similarity, generation diversity, and privacy leakage. We also evaluate these models on our novel graph similarity and DCR ratio metrics.

The results indicate that TabDDPM achieves the overall best performance across various metrics. However, it appears that TabDDPM may struggle to handle categorical variables with hundreds of classes. Additionally, the findings reveal the performance gap of the current generative models, and the prevalence of mode collapse, underscoring the need to develop models specifically tailored to domains such as transportation.

Furthermore, extending the evaluation beyond the New York City taxi data is expected to offer more insights on the current tabular generative models. 

% \section*{Acknowledgments}
% The research reported herein was supported in part by ?

\begin{credits}
 % \subsection{\ackname}The research reported herein was supported in part by ?.
\end{credits}

\addcontentsline{toc}{section}{References} % add references to toc
\bibliographystyle{splncs04}
\bibliography{references}

\end{document}